\documentclass[conference, a4paper]{IEEEtran}

\usepackage{comment}
\usepackage{color}

\usepackage{algorithm}
\usepackage{algpseudocode}
\usepackage{amsmath, amssymb}
\usepackage{multirow}
\usepackage{bbm}
\usepackage[switch,columnwise]{lineno}

\DeclareMathOperator*{\myargmin}{arg\,min}
\DeclareMathOperator*{\myargmax}{arg\,max}

\ifCLASSINFOpdf
   \usepackage[pdftex]{graphicx,xcolor}
   \usepackage[colorlinks,citecolor=green,urlcolor=blue,bookmarks=false,hypertexnames=true]{hyperref} 
   \usepackage{subfigure}
\else
\fi
\begin{document}
%
\title{PMSSC: Parallelizable multi-subset based self-expressive model for subspace clustering}

\author{\IEEEauthorblockN{Katsuya Hotta}
\IEEEauthorblockA{Iwate University\\
Iwate, 020-8551 Japan\\
Email: hotta@iwate-u.ac.jp}

\and

\IEEEauthorblockN{Takuya Akashi}
\IEEEauthorblockA{Iwate University\\
Iwate, 020-8551, Japan\\
Email: akashi@iwate-u.ac.jp }

\and

\IEEEauthorblockN{Shogo Tokai}
\IEEEauthorblockA{University of Fukui \\
Fukui, 910-8507 Japan\\
Email: 	tokai@u-fukui.ac.jp}

\and

\IEEEauthorblockN{Chao Zhang}
\IEEEauthorblockA{University of Fukui \\
Fukui, 910-8507 Japan\\
Email: zhang@u-fukui.ac.jp}
}


%




\maketitle
\thispagestyle{plain}
\pagestyle{plain}
\begin{abstract}
Subspace clustering methods which embrace a self-expressive model that represents each data point as a linear combination of other data points in the dataset provide powerful unsupervised learning techniques. However, when dealing with large datasets, representation of each data point by referring to all data points via a dictionary suffers from high computational complexity.
To alleviate this issue, we introduce a parallelizable multi-subset based self-expressive model (PMS) which represents each data point by combining multiple subsets, with each consisting of only a small proportion of the samples. The adoption of PMS in subspace clustering (PMSSC) leads to computational advantages because the optimization problems decomposed over each subset are small, and can be solved efficiently in parallel.
Furthermore, PMSSC is able to combine multiple self-expressive coefficient vectors obtained from subsets, which contributes to an improvement in self-expressiveness.
Extensive experiments on synthetic and real-world datasets show the efficiency and effectiveness of our approach in comparison to other methods.

\end{abstract}


%
\IEEEpeerreviewmaketitle

\section{Introduction}
\label{sec:sec1}
In many real-world cases, approximating high-dimensional data as a union of low-dimensional subspaces is a beneficial technique for reducing computational complexity and the effects of noise. The task of subspace clustering~\cite{vidal2011subspace, hotta2021affine}, which is the segmentation of a set of data points into those lying on certain subspaces, has been studied in many practical applications such as face clustering~\cite{zhang2019energy}, image segmentation~\cite{yang2008unsupervised}, motion segmentation~\cite{vidal2008multiframe}, scene segmentation~\cite{tierney2014subspace}, and homography detection~\cite{zhang2020g2mf}.
Recently, self-expressive models~\cite{vidal2009sparse, elhamifar2013sparse} have been explored, which embrace the self-expressive property of subspaces to compute an affinity matrix.
The self-expressive property states that each data point from a union of subspaces can be represented as a linear combination of other points.
Specifically, given a data matrix $X \in \mathbb{R}^{D \times N}$ in which each data point is a column, the self-expressive model of data point $\boldsymbol{x}_{i} \in \mathbb{R}^{D}$ can be described as
\begin{equation}
    \boldsymbol{x}_{i} = X\boldsymbol{c}_{i}, \quad c_{ii}=0, 
    \label{equ:equ1}
\end{equation}
where $\boldsymbol{c}_{i} \in \mathbb{R}^{N}$ is a coefficient vector, and the constraint $c_{ii} = 0$ avoids the trivial solution of representing a point as a linear combination of itself.
The feasible solutions of Eq. (\ref{equ:equ1}) are generally not unique because the number of data points lying on a subspace is larger than its dimensionality.
However, at least one $\boldsymbol{c}_{i}$ exists where $c_{ij}$ is nonzero only if data points $\boldsymbol{x}_{i},\boldsymbol{x}_{j}$ are in the same subspace, and such a state is called subspace-preserving~\cite{you2016scalable}.
Previous works have tried to compute subspace-preserving representations by imposing a regularization term on the coefficients $\boldsymbol{c}_{i}$.
In particular, one  algorithm for obtaining a sparse solution to Eq. (\ref{equ:equ1}), sparse subspace clustering (SSC)~\cite{vidal2009sparse, elhamifar2013sparse}, can recover subspaces under mild conditions by regularizing the coefficient matrix $C := [\boldsymbol{c}_{1}, \ldots, \boldsymbol{c}_{N}] \in \mathbb{R}^{N \times N}$ corresponding to the coefficient vector of each data point $\boldsymbol{x}_{i}$.
SSC not only achieves high clustering accuracy for datasets with outliers and missing entries, but also has the useful properties of giving theoretical guarantees and providing modeling flexibility, which have influenced many other approaches such as~\cite{guo2021efficient, you2020self}.
However, SSC suffers from high computational and memory costs when dealing with a large-scale dataset because of the need to determine the $\mathcal{O}(N^{2})$ coefficients of $C$. In light of these problems, there has been much interest in recent years in developing scalable subspace clustering algorithms that can be applied to large-scale datasets, taking advantage of the ease of analyzing computational complexity due to the simplicity of the model.

\begin{figure*}[!t]
\centering
    \includegraphics[width=\linewidth]{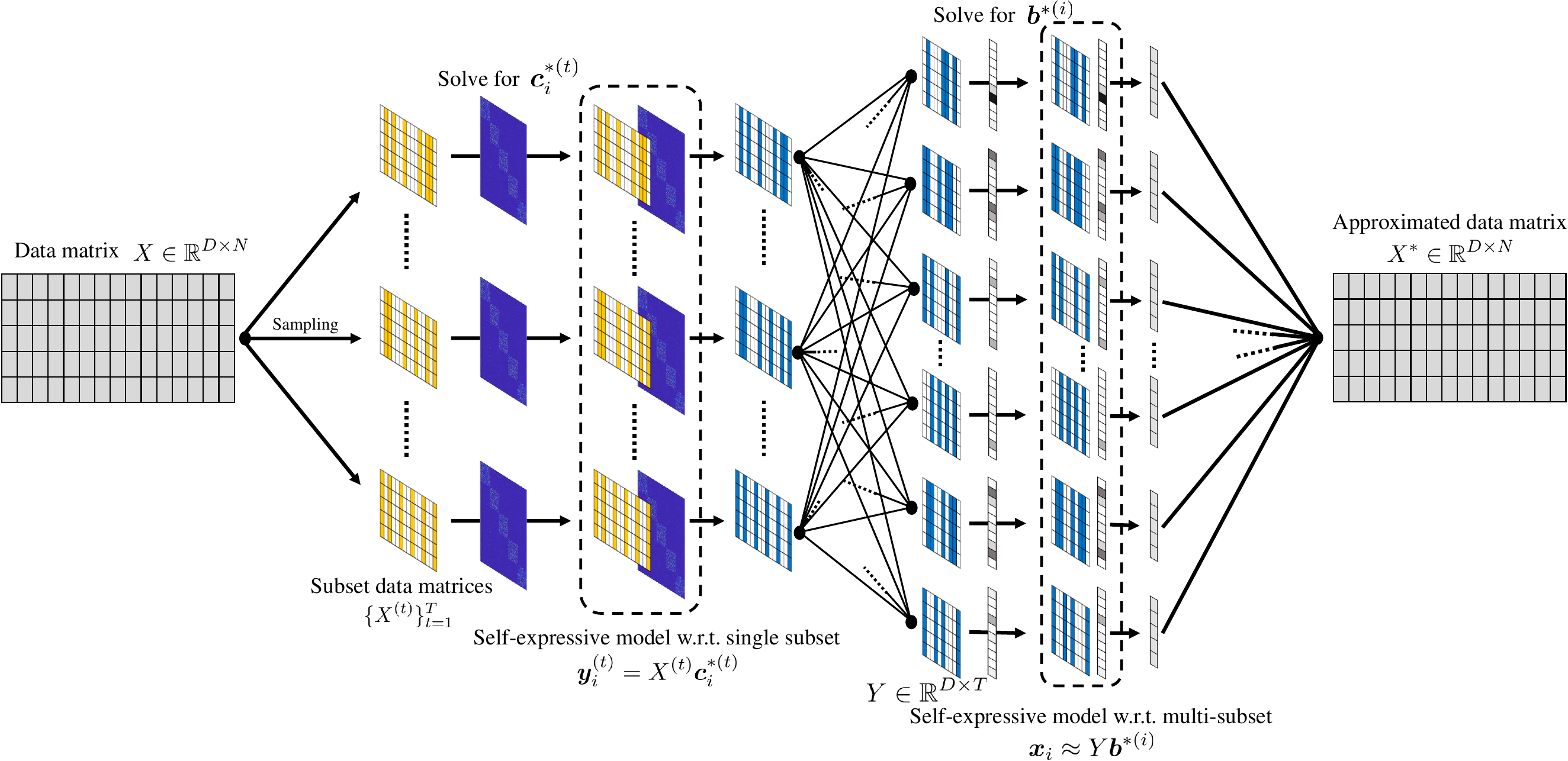}
  \caption{Overview of our self-expressive model. Given a data matrix $X$ in which each data point $\boldsymbol{x}_{i}$ is a column, our approach represents a self-expressive model over the entire data by combining multiple subsets generated by sampling (Algorithm~\ref{alg:alg1}). Specifically, our method computes the self-expressive data point $\boldsymbol{y}_{i}^{(t)}$ by solving for the self-expressive coefficient vector $\boldsymbol{c}_{i}^{\ast(t)}$ for each point $\boldsymbol{x}_{i}$ in $T$ subsets (Algorithm~\ref{alg:alg2}). Then, the self-expressive properties of the entire data are obtained by solving for $\boldsymbol{b}^{\ast}$ using $Y$ with each data point $\boldsymbol{y}_{i}^{(t)}$ computed from each subset as columns (Algorithm~\ref{alg:alg3}).
  }
  \label{fig:fig1}
\end{figure*}
Several works have attempted to address the problem of computational cost for large-scale datasets using a sampling strategy,  motivated by the sparsity assumption that each data point can be represented as a linear combination of a few basis vectors. The self-expressive property with a few sampled data points and classifying of the other data points was proposed in~\cite{peng2013scalable}.
While this strategy can produce clustering results more efficiently for a large-scale dataset than directly applying SSC to all data, it leads to poor clustering performance when the sampled data is not representative of the original dataset. 
Although a learning-based sampling method has also been proposed for generating a coefficient matrix that is representative of the original dataset~\cite{matsushima2019selective}, the accuracy and computational complexity still depend largely on the size of the subsets, as these methods attempt to solve for a self-expressive model in a single subset. 
Also, no effort has been made to explicitly improve the self-expressiveness of the self-expressive coefficient vectors in these methods.

To further improve self-expressiveness without increasing the computational burden, in this paper, we propose a self-expressive model adopting multiple subsets, which is computable in parallel.
Specifically, our model obtains a self-expressive coefficient matrix by combining multiple subsets; each subset consists of only a small proportion of the samples.
This strategy not only enjoys the benefit of low computational cost like other single subset-based methods, but also is more effectively subspace-preserving because the representation of the original data is a linear combination of multiple self-expressive coefficient vectors.

Our contributions are highlighted as follows:
\begin{itemize}
    \item  a novel clustering approach that exploits a self-expressive model based on multiple subsets,
    \item a concisely formulated model, 
    \item each subset can be computed independently in parallel without additional computational overhead,
    \item extensive experiments on both synthetic data and real-world datasets showing that our proposed method can achieve better results without increasing processing time.
\end{itemize}

\section{Related Work}
\subsection{Background}
\label{sec:sec2}
In the past few years, there has been a surge of spectral clustering-based algorithms that segment a set of data points by performing spectral clustering. Previous classical methods, such as $k$-subspaces~\cite{tseng2000nearest} and median $k$-flats~\cite{zhang2009median}, assume that the dimensionalities of the underlying subspaces are given in advance. This latent knowledge is generally hard to access in many real-world applications. In addition, these methods are usually non-convex and thus sensitive to initialization~\cite{lipor2021subspace, lane2019adaptive}. 
Aiming to relax the limitations of the $k$-subspace algorithm, the majority of modern subspace clustering methods explored have turned to spectral clustering~\cite{shi2000normalized, von2007tutorial}, which segment data using an affinity matrix that captures whether a certain pair of data points lie on the same subspace. While many early  methods~\cite{lu2018subspace, dong2019sparse, you2020self, hotta2020imvip} achieve better segmentation than classical clustering algorithms even without the latent knowledge, these methods produce erroneous segmentation results for data points near the intersection of two subspaces due to the dense sampling of points lying on the subspace~\cite{yan2006general}. 
We now introduce previous subspace clustering approaches based on spectral clustering, then describe various techniques of scalable subspace clustering methods for dealing with large-scale datasets, which are closer to our proposed method.

\subsection{Subspace Clustering Using Spectral Clustering}
Most subspace clustering approaches based on spectral clustering consist of two phases: (i) computing an affinity matrix based on the nonzero coefficients that appear in the representation of each data point as a combination of other points, and (ii) segmenting data points from the computed affinity matrix by applying spectral clustering.
The key to the success of segmentation is the phase of computing the affinity matrix.
Therefore, many methods have been proposed to compute the affinity matrix.
For example, local subspace affinity~\cite{yan2006general} and spectral curvature clustering (SCC)~\cite{chen2009spectral} find neighborhoods based on the observation that a point and its $k$-nearest neighbors often lie on the same subspace.
However, the computational complexity of finding multi-way similarity in these methods grows exponentially with the number of subspace dimensions, motivating the use of a sampling strategy to lower the computational complexity~\cite{elhamifar2013sparse}.
Recently, the self-expressive model, which employs the self-expressive property in Eq. (\ref{equ:equ1}), has become the most popular one.
In particular, SSC takes advantage of sparsity~\cite{donoho2006most} by adopting $\ell_{1}$ norm regularization of the coefficient vector to achieve high clustering performance.
This idea has motivated many methods, using the $\ell_{2}$ norm in least squares regression~\cite{lu2012robust}, the nuclear norm in low rank representation (LRR)~\cite{liu2010robust}, the $\ell_{1}$ plus $\ell_{2}$ norm in elastic net subspace clustering (EnSC)~\cite{you2016oracle}, and the Frobenius norm in efficient dense subspace clustering~\cite{ji2014efficient}.
In practice, however, solving the $\ell_{p}$ norm minimization problem for large-scale data may be prohibitive.
Also, the memory required becomes larger as the amount of data increases.

\subsection{Scalable Subspace Clustering}
When constructing the affinity matrix, several methods based on spectral clustering suffer from high computational complexity. To reduce the computational complexity of this phase, a sparse self-expressive model adopting a greedy algorithm was proposed in~\cite{dyer2013greedy, you2016scalable}.
However, these approaches lead to unsatisfactory clustering results if the nonzero elements do not contain sufficient connections within each optimized coefficient vector~\cite{nasihatkon2011graph}.
Other popular approaches to alleviate the computational and memory loads were inspired by a sampling strategy.
In~\cite{peng2013scalable}, scalable sparse subspace clustering (SSSC) is computationally efficient, using a subset generated by random sampling.
However, because the random sampling method relies on a single subset, data points from the same subspace will not be represented by the self-expressive model if they are not appropriately sampled. Exemplar-based subspace clustering~\cite{you2018scalable, you2020self} is an efficient sampling technique that iteratively selects the least well-represented point as a subset to address the problem.
Selective sampling-based scalable sparse subspace clustering (S$^5$C)~\cite{matsushima2019selective}, which generates a subset by selective sampling, provides approximation guarantees of the subspace-preserving property.
In~\cite{chen2020stochastic}, the subspace-preserving representations are found by solving a consensus problem with multiple subsets to improve the connectivity of the affinity matrix.
In~\cite{You2016-jl}, a divided-and-conquer framework using multiple subsets obtained by separating the entire dataset is proposed.
While this approach can deal with large-scale data, final segmentation results depend on the self-expressive properties of the optimized self-expressive coefficient vectors of each subset.
Our method differs significantly from~\cite{You2016-jl} and~\cite{chen2020stochastic} in that our proposed self-representation model is designed to minimize the difference from the original data points by combining the self-expressive property of multiple subsets.
Lastly, in this paper, we limit our discussion to non-deep learning approaches, which are more mathematically straightforward to explain and rely less on parameter tuning.

\section{Parallelizable Multi-Subset Based Sparse Subspace Clustering}
\label{sec:sec3}
\subsection{Problem and Approach}
As a problem definition, our final goal is to find the self-expressive coefficient vector $\boldsymbol{c}_{i}$, which satisfies the subspace-preserving representation in Eq. (\ref{equ:equ1}).
That is, the self-expressive residual can be obtained by solving the following optimization problem,
\begin{equation}
    \min_{\boldsymbol{c}_{i}}\| \boldsymbol{x}_{i} - X \boldsymbol{c}_{i}\|_{2}^{2} \quad \mathrm{such~that} \quad \| \boldsymbol{c}_{i}\|_{0} \leq s,\  c_{ii}=0,
    \label{equ:equ2}
\end{equation}
where $\|\cdot\|_{0}$ is the $\ell_{0}$ pseudo-norm that returns the number of nonzero entries in the vector. 
This optimization problem has been shown~\cite{davenport2010analysis, tropp2004greed} to recover provably subspace-preserving solutions using the orthogonal matching pursuit (OMP) algorithm~\cite{pati1993orthogonal}.
$s$ is a tuning parameter for the OMP algorithm, which controls the sparsity of the solution by selecting up to $s$ entries in the coefficient vector $\boldsymbol{c}_{i}$.
Although the OMP algorithm is computationally efficient and is guaranteed to give subspace-preserving solutions under mild conditions, it is unable to produce a subspace-preserving solution with a number of nonzero entries  exceeding the dimensionality of the subspace~\cite{you2016scalable}.
This leads to poor clustering performance with too sparse affinity between data points, especially when the density of data points lying on the subspace is low.

We propose a novel subspace clustering algorithm with a parallelizable multi-subset based self-expressive model, as illustrated in Fig.~\ref{fig:fig1}. Sec.~\ref{sec:sec3_1} introduces our proposed self-expressive model that extends the model in Eq. (\ref{equ:equ2}) to multiple subsets via a sampling technique.
Sec.~\ref{sec:sec3_2} then explains the solution of our self-expressive model by the OMP algorithm.
Finally, we summarize the proposed subspace clustering algorithm in Algorithm~\ref{alg:alg4}.

\subsection{Parallelizable Multi-Subset based Self-Expressive Model}
\label{sec:sec3_1}
To deal with large-scale data, we first generate $T$ index subsets from the whole dataset by weighted random sampling~\cite{wong1980efficient} as follows:
\begin{equation}
    \mathcal{I}^{(t)} \subset [N] \quad \mathrm{s.t.} \quad n(\mathcal{I}^{(t)}) = \lceil \delta N \rceil,\ t=1,\ldots,T,
    \label{equ:equ3}
\end{equation}
where $\mathcal{I}^{(t)}$ is the index set of the $t$-th subset that is sampled with probability proportional to the elements of the weight vector $\boldsymbol{w}^{(t)} \in \mathbb{R}^{N}$, $[N]$ is $N$ indices $\{1,\ldots,N\}$, $0 < \delta \leq 1$ is the sampling rate, and $n(\cdot)$ is the cardinality function that is a measure of the number of elements.
The $t$-th selected element of $\boldsymbol{w}^{(t)}$ is updated as $\boldsymbol{w}_{i}^{(t+1)} = 0.1\boldsymbol{w}_{i}^{(t)}$.
Then, in each sampled $t$-th subset, the optimization problem in Eq. (\ref{equ:equ2}) can be expressed as follows:
\begin{equation}
\begin{split}
    \boldsymbol{c}_{i}^{\ast(t)} = &
    \myargmin_{\boldsymbol{c}_{i}^{(t)}}\| \boldsymbol{x}_{i}^{(t)}-X^{(t)}\boldsymbol{c}_{i}^{(t)}\|_{2}^{2} \\
  &\mathrm{s.t.} \hspace{3px} \| \boldsymbol{c}_{i}^{(t)}\|_{0} \leq s,\quad c_{ii}^{(t)}=0,
\end{split}
\label{equ:equ4}
\end{equation}
where $X^{(t)}\in\mathbb{R}^{D \times N}$ is the data matrix of the randomly sampled $t$-th subset.
$\boldsymbol{c}_{i}^{(t)} \in \mathbb{R}^{N}$ is the self-expressive coefficient vector for each data point $\boldsymbol{x}_{i}^{(t)}$ in the $t$-th subset.
Note that to ensure the dimensionality of $\boldsymbol{c}_{i}^{(t)}$ is $N$, the columns of each data matrix $X^{(t)}$ corresponding to the non-sampled indices are replaced by zero-vectors: $\boldsymbol{x}_{i}^{(t)}=\boldsymbol{0},\  \forall i\notin \mathcal{I}^{(t)}$.
From each optimized coefficient vector $\boldsymbol{c}_{i}^{\ast(t)}$, each data point $\boldsymbol{x}_{i}^{(t)}$ can be represented by a self-expressive model,  given by:
\begin{equation}
    \boldsymbol{y}_{i}^{(t)}=X^{(t)} \boldsymbol{c}_{i}^{\ast(t)} \quad \mathrm{s.t.} \quad c_{ii}^{\ast(t)}=0,
    \label{equ:equ5}
\end{equation}
\begin{algorithm}[t]
\caption{Optimization for the parallelizable multi-subset based self-expressive model (PMS)}
\label{alg:alg1}  
\begin{algorithmic}[1]
\renewcommand{\algorithmicrequire}{\textbf{Input:}}
\renewcommand{\algorithmicensure}{\textbf{Output:}}
  \Require{Data matrix $X \in \mathbb{R}^{D \times N}$,  number of subsets $T$, sampling rate $\delta$, maximum number of repetitions $s$,  error term $\epsilon$}
  \State{Generate $T$ index subsets $\{\mathcal{I}^{(t)}\}_{t=1}^{T}$ via Eq. (\ref{equ:equ3});}
  \State{Generate $T$ subset data matrices $\{X^{(t)}\}_{t=1}^{T}$ based on $\{\mathcal{I}^{(t)}\}_{t=1}^{T}$;}
  \For{$i=1,\ldots,N$}
  \For{$t=1,\ldots,T$}
  \State{Given $X^{(t)}$ and $\boldsymbol{x}_{i}^{(t)}$, solve for $\boldsymbol{c}_{i}^{\ast(t)}$ via Algorithm~\ref{alg:alg2};}
  \State{Given $\boldsymbol{c}_{i}^{\ast(t)}$, compute $\boldsymbol{y}_{i}^{(t)}$ for all data points via Eq. (\ref{equ:equ5});}
 \EndFor{}
  \State{Set $Y=[\boldsymbol{y}_{i}^{(1)},\ldots,\boldsymbol{y}_{i}^{(T)}] \in \mathbb{R}^{D \times T}$;}
  \State{Given $Y$ and $\boldsymbol{x}_{i}$, solve for $\boldsymbol{b}^{\ast(i)}$ via Algorithm~\ref{alg:alg3};}
  \State{Given $\boldsymbol{c}_{i}^{\ast(t)}$ and $\boldsymbol{b}^{\ast(i)}$, compute $\boldsymbol{c}_{i}^{\ast}$ via Eq. (\ref{equ:equ10});}
  \EndFor {}
  \State{Set $C^{\ast}=[\boldsymbol{c}_{1}^{\ast},\ldots,\boldsymbol{c}_{N}^{\ast}] \in \mathbb{R}^{N\times N}$;}
  \Ensure{Coefficient matrix $C^{\ast}$}
\end{algorithmic}
\end{algorithm}
where $\boldsymbol{y}_{i}^{(t)}$ is the data point computed by the self-expressive model from the $t$-th subset.
In practice, however, the data point $\boldsymbol{y}_{i}^{(t)}$ in Eq. (\ref{equ:equ5}) generally has an error term $\boldsymbol{z}_{i}$, i.e., $\boldsymbol{y}_{i}^{(t)} = \boldsymbol{x}_{i} +\boldsymbol{z}_{i}$, because of the limitations of  using $X^{(t)}$ as a dictionary for reconstruction.
To minimize $\boldsymbol{z}_{i}$, we first represent $\boldsymbol{x}_{i}$ as a linear combination of $\boldsymbol{y}_{i}^{(t)}$, as follows:
\begin{equation}
\begin{split}
    \boldsymbol{x}_{i} &\approx \sum_{t=1}^{T} b_{t}^{(i)} (X^{(t)} \boldsymbol{c}_{i}^{\ast(t)})\\
    &\approx \sum_{t=1}^{T} b_{t}^{(i)}\boldsymbol{y}_{i}^{(t)},
\end{split}
    \label{equ:equ6}
\end{equation}
where $\boldsymbol{b}^{(i)} \in \mathbb{R}^{T}$ is the weight coefficient vector to represent $\boldsymbol{x}_{i}$, and $b_{t}^{(i)} \in \mathbb{R}$ is the $t$-th entry of $\boldsymbol{b}^{(i)}$.
The coefficient vector $\boldsymbol{b}^{(i)}$ of the linear combination in Eq. (\ref{equ:equ6}) can be obtained by solving the following optimization problem,
\begin{equation}
    \boldsymbol{b}^{\ast (i)} = \myargmin_{\boldsymbol{b}^{(i)}} \| \boldsymbol{x}_{i} - \sum_{t=1}^{T} b_{t}^{(i)} \boldsymbol{y}_{i}^{(t)}\|_{2}^{2}.
    \label{equ:equ7}
\end{equation}
For simplicity, we introduce a data matrix $Y=[\boldsymbol{y}_{i}^{(1)},\ldots,\boldsymbol{y}_{i}^{(T)}] \in \mathbb{R}^{D \times T}$ with each data point $\boldsymbol{y}_{i}^{(t)}$ from Eq. (\ref{equ:equ5}) as  columns, and rewrite Eq. (\ref{equ:equ7}) as
\begin{equation}
    \boldsymbol{b}^{\ast(i)} = \myargmin_{\boldsymbol{b}^{(i)}} \| \boldsymbol{x}_{i} - Y \boldsymbol{b}^{(i)}\|_{2}^{2}.
    \label{equ:equ8}
\end{equation}
This is the formulation of the optimization problem for subspace clustering in Eq. (\ref{equ:equ2}), and can be further described as:
\begin{equation}
    \boldsymbol{x}_{i} \approx Y \boldsymbol{b}^{\ast(i)}.
    \label{equ:equ9}
\end{equation}
Unlike in Eq. (\ref{equ:equ1}), here $Y$ is the data matrix computed from each subset to represent $\boldsymbol{x}_{i}$.
Thus, no constraint is required to avoid the trivial solution of representing a point as a linear combination of itself.
To explicitly express Eq. (\ref{equ:equ2}), the self-expressive coefficient vector $\boldsymbol{c}_{i}^{\ast}$ corresponding to $X$ is obtained by
\begin{equation}
    \boldsymbol{c}_{i}^{\ast} = \sum_{t=1}^{T}b_{t}^{\ast (i)} \boldsymbol{c}_{i}^{\ast (t)}.
    \label{equ:equ10}
\end{equation}
It is worth noting that each $\boldsymbol{c}^{\ast{(t)}}$ can be determined independently from each subset, so can be computed in parallel for speed.

\begin{algorithm}[t]
\caption{OMP algorithm for finding $\boldsymbol{c}_{i}^{\ast(t)}$}     
\label{alg:alg2}  
\begin{algorithmic}[1]
\renewcommand{\algorithmicrequire}{\textbf{Input:}}
\renewcommand{\algorithmicensure}{\textbf{Output:}}
\Require{Data matrix $X^{(t)} \in \mathbb{R}^{D \times N}$, reference data point $\boldsymbol{x}_{i}^{(t)}$, maximum repetition count $s$,  error term $\epsilon$}
\State{Initialize $k=0$, residual $\boldsymbol{r} = \boldsymbol{x}_{i}^{(t)}$,  support set $\mathcal{S}=\emptyset$;}
\While{$k < s$ and $\|\boldsymbol{r}\|_{2} > \epsilon$}
\State{Find $j^{\ast}$ via Eq. (\ref{equ:equ11});}
\State{$\mathcal{S} \leftarrow \mathcal{S} \cup \{j^{\ast}\}$;}
\State{Estimate $\boldsymbol{c}_{i}^{\ast(t)}$ via Eq. (\ref{equ:equ12});}
\State{Update the residual $\boldsymbol{r}$ via Eq. (\ref{equ:equ13});}
\State{$k = k + 1$;}
\EndWhile{}
\Ensure{Self-expressive coefficient vector $\boldsymbol{c}_{i}^{\ast(t)}$}
\end{algorithmic}
\end{algorithm}

\begin{algorithm}[t]
\caption{OMP algorithm for finding $\boldsymbol{b}^{\ast(i)}$} 
\label{alg:alg3}  
\begin{algorithmic}[1]
\renewcommand{\algorithmicrequire}{\textbf{Input:}}
\renewcommand{\algorithmicensure}{\textbf{Output:}}
  \Require{Data matrix $Y \in \mathbb{R}^{D \times T}$, reference data point $\boldsymbol{x}_{i}$,  error term $\epsilon$}
  \State{Initialize $k=0$, residual $\boldsymbol{r} = \boldsymbol{x}_{i}$,  support set $\mathcal{S}=\emptyset$;}
 \While{$k < T$ and $\|\boldsymbol{r}\|_{2} > \epsilon$}
  \State{Find $j^{\ast}$ via Eq. (\ref{equ:equ14});}
  \State{Update $\mathcal{S} \leftarrow \mathcal{S} \cup \{j^{\ast}\}$;}
  \State{Estimate $\boldsymbol{b}^{\ast(i)}$ via Eq. (\ref{equ:equ15});}
  \State{Update the residual $\boldsymbol{r}$ via Eq. (\ref{equ:equ16});}
  \State{$k = k + 1$;}
  \EndWhile{}
  \Ensure{Weight coefficient vector $\boldsymbol{b}^{\ast(i)}$}
\end{algorithmic}
\end{algorithm}

\subsection{Optimization with Orthogonal Matching Pursuit}
\label{sec:sec3_2}
In this section, we show that the parameters of the proposed PMS model can be determined by dividing the optimization problem into two small optimization problems as summarized in Algorithm~\ref{alg:alg1}.
Overall, Eq. (\ref{equ:equ10}) can be determined by solving the minimization problems in Eqs. (\ref{equ:equ4}) and (\ref{equ:equ8}). We  introduce both of the minimization procedures below.
Specifically, initially, $T$ subset data matrices $\{X^{(t)}\}_{t=1}^{T}$ are generated based on the sampling set $\mathcal{I}^{(t)}$ in Eq. (\ref{equ:equ3}).

To efficiently solve for $\boldsymbol{c}_{i}^{\ast(t)}$ in Eq. (\ref{equ:equ4}), we introduce Algorithm~\ref{alg:alg2} based on the OMP algorithm.
The support set and the residual are initialized to $\mathcal{S} = \emptyset$ and  $\boldsymbol{r} = \boldsymbol{x}_{i}^{(t)}$, respectively.
$\mathcal{S}$ denotes the index set, which is updated on each iteration by adding one index $j^{\ast}$. $j^{\ast}$ is computed using
\begin{equation}
    j^{\ast} = \myargmax_{j \in \mathcal{I}^{(t)} \setminus \mathcal{S}} \boldsymbol{x}_{j}^{(t)\top}\boldsymbol{r}.
    \label{equ:equ11}
\end{equation}
Then, using the updated $\mathcal{S}$, the self-expressive coefficient vector $\boldsymbol{c}_{i}^{\ast(t)}$ is found by solving the following problem:
\begin{eqnarray}
    &\hspace{-3mm}\boldsymbol{c}_{i}^{\ast(t)} = 
    \begin{cases}
     \myargmin_{\boldsymbol{c}_{i}^{(t)}} \| \boldsymbol{x}_{i}^{(t)}-X^{(t)}\boldsymbol{c}_{i}^{(t)}\|_{2}^{2}, & \mathrm{if} \hspace{2pt} i\in \mathcal{I}^{(t)}, \\
     \boldsymbol{0}, & \mathrm{otherwise},
  \end{cases}\nonumber\\
  &\mathrm{such~that} \quad \mathrm{supp}(\boldsymbol{c}_{i}^{(t)}) \subseteq \mathcal{S},
      \label{equ:equ12}
\end{eqnarray}
where $\mathrm{supp}(\boldsymbol{c}_{i}^{(t)})$ is the support function that returns the subgroup of the domain containing the elements not mapped to zero.
$\boldsymbol{r}$ is updated using:
\begin{equation}
    \boldsymbol{r} \leftarrow \boldsymbol{x}_{i}^{(t)} - X^{(t)} \boldsymbol{c}_{i}^{\ast(t)}.
    \label{equ:equ13}
\end{equation}
This process is repeated until the number of iterations $k$ reaches its limit $s$ or $\boldsymbol{r}$ is smaller than the error $\epsilon$.

\begin{algorithm}[t]
\caption{Parallelizable multi-subset based sparse subspace clustering (PMSSC)}       
\label{alg:alg4}  
\begin{algorithmic}[1]
\renewcommand{\algorithmicrequire}{\textbf{Input:}}
\renewcommand{\algorithmicensure}{\textbf{Output:}}
  \Require{Data matrix $X=[\boldsymbol{x}_{1},\ldots,\boldsymbol{x}_{N}]  \in \mathbb{R}^{D \times N}$, parameters $T$, $\delta$, $s$,  $\epsilon$}
  \State{Compute coefficient matrix $C^{\ast}$ via Algorithm~\ref{alg:alg1};}
  \State{Define affinity matrix  $W = |C^{\ast}| + |C^{\ast T}|$;}
  \State{Apply spectral clustering;}
  \Ensure{Clustering results of $X$}
\end{algorithmic}
\end{algorithm}

To find $\boldsymbol{b}^{\ast(i)}$,
Eq. (\ref{equ:equ8}) can also be solved via the OMP algorithm, as shown in Algorithm~\ref{alg:alg3}.
The input data matrix $Y \in \mathbb{R}^{D \times T}$ is generated by Eq. (\ref{equ:equ5}) from $\boldsymbol{c}_{i}^{\ast(t)}$.
Note that the size of $Y$ depends on the number of subsets $T$ and  is much smaller than the number of data points. The maximum number of repetitions  is $T$, and $\mathcal{S}$ is updated by finding the index $j^{\ast}$ satisying
\begin{equation}
    j^{\ast} = \myargmax_{j\in [T] \setminus \mathcal{S}} \boldsymbol{y}_{i}^{(j)\top}\boldsymbol{r}.
    \label{equ:equ14}
\end{equation}
In addition,  the weight coefficient vector $\boldsymbol{b}^{\ast(i)}$ and  update of $\boldsymbol{r}$ are determined by solving
\begin{equation}
     \boldsymbol{b}^{\ast(i)} = \myargmin_{\boldsymbol{b}^{(i)}} \| \boldsymbol{x}_{i} - Y \boldsymbol{b}^{(i)}\|_{2}^{2}, \hspace{10px} \mathrm{s.t.} \hspace{3px} \mathrm{supp}(\boldsymbol{b}_{i}) \subseteq \mathcal{S}.
    \label{equ:equ15}
\end{equation}
\begin{equation}
    \boldsymbol{r} \leftarrow \boldsymbol{x}_{i} - Y \boldsymbol{b}_{t}^{\ast(i)}.
    \label{equ:equ16}
\end{equation}

For clarity, we summarize the entire framework of our proposed subspace clustering approach in Algorithm~\ref{alg:alg4}, calling it the \emph{parallelizable multi-subset based sparse subspace clustering} (PMSSC) method. Given $X$ and parameters $T$, $\delta$, $s$, and $\epsilon$, the optimal solution $C^{\ast}$ can be found using Algorithm~\ref{alg:alg1}.
We thus define the affinity matrix as $A = |C^{\ast}| + |C^{\ast T}|$ using the computed $C^{\ast}$; the final clustering results can be obtained by applying spectral clustering to $A$ via normalized cut~\cite{shi2000normalized}.

\section{Experiments and Results}
\label{sec:sec4}
We have evaluated  our approach using both synthetic data and real-world benchmark datasets. 

\subsection{Baselines and Evaluation Metrics.}
We compare our approach to the following eight methods: SCC~\cite{chen2009spectral}, LRR~\cite{liu2010robust}, thresholding-based subspace clustering (TSC)~\cite{heckel2013subspace}, low rank subspace clustering (LRSC)~\cite{vidal2014low}, SSSC~\cite{peng2013scalable}, EnSC~\cite{you2016oracle}, SSC-OMP~\cite{you2016scalable}, and S$^5$C~\cite{matsushima2019selective}.
Tests for all comparative methods used provided source code, and each parameter was carefully tuned to give the best clustering accuracy.
For spectral clustering, except for SCC, S$^{5}$C, and SSSC, we applied  normalized cut~\cite{shi2000normalized} to the affinity matrix $A = |C| + |C^{T}|$. (SCC and S$^{5}$C have their own spectral clustering phase, while SSSC obtains  clustering results from the data split into two parts). 
Unlike SSC-OMP, our method, which involves independent calculation for each subset, can be implemented in parallel with multi-core processing.
All algorithms ran on an AMD Ryzen 7 3700x processor with 32 GB RAM.
Following~\cite{you2016scalable}, as  quantitative evaluation metrics, we evaluated each algorithm using clustering accuracy (acc: $a\%$), subspace-preserving representation error (sre: $e\%$), connectivity (conn: $c$), and runtime ($t$ seconds).
Clustering accuracy represents the percentage of correctly labeled data points:
\begin{equation}
    a = \max_{\pi} \frac{100}{N}\sum_{ij}Q_{\pi(i)j}^\mathrm{est}Q_{ij}^\mathrm{true},
    \label{equ:acc}
\end{equation}
where $\pi$ is a permutation of the $L$ cluster groups.
$Q^\mathrm{est}$ and $Q^\mathrm{true}$ are the estimated labeling result and ground-truth, respectively, scoring one in the $(i,j)$th entry if a data point $j$ belongs to the $i$-th cluster and zero otherwise.
The subspace-preserving representation error indicates the average fraction of affinities formed from other subspaces in each $\boldsymbol{c}_{j}$,
\begin{equation}
    e = \frac{100}{N} \sum_{j}\left(1-\sum_{i}(\omega_{ij}|c_{ij}|)/\|\boldsymbol{c}_{j}\|_{1}\right),
    \label{equ:sre}
\end{equation}
where $\omega_{ij} \in \{0,1\}$ is the true affinity, and $\| \cdot \|_{1}$ returns the $\ell_{1}$ norm.
The connectivity shows the average connection of the affinity matrix with $L$ cluster groups as follows:
\begin{equation}
    c = \frac{1}{L} \sum_{i=1}^{L} \lambda_{2}^{(i)},
    \label{equ:conn}
\end{equation}
where $\lambda_{2}$ is the second smallest eigenvalue of the normalized Laplacian for each of the $L$ subgraph, and $\lambda_{2}^{(i)}$ indicates the algebraic connectivity for each cluster.
If $c= 0$, then  at least one subgraph  is not connected~\cite{chung1997spectral}. 

\begin{figure*}[t]
    \centering
    \subfigure[]{\includegraphics[width=0.35\linewidth]{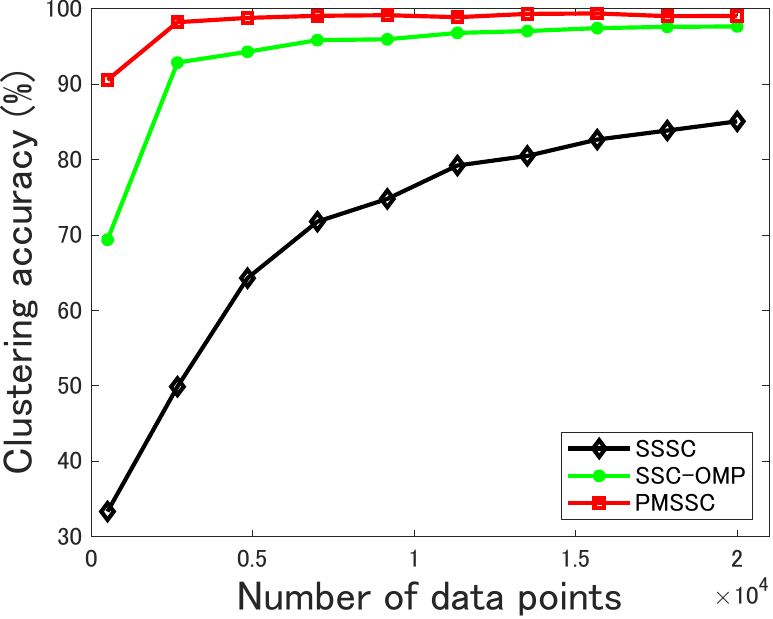}
    \label{fig:res_syn_a}
    }\hspace{10pt}
    \subfigure[]{\includegraphics[width=0.354\linewidth]{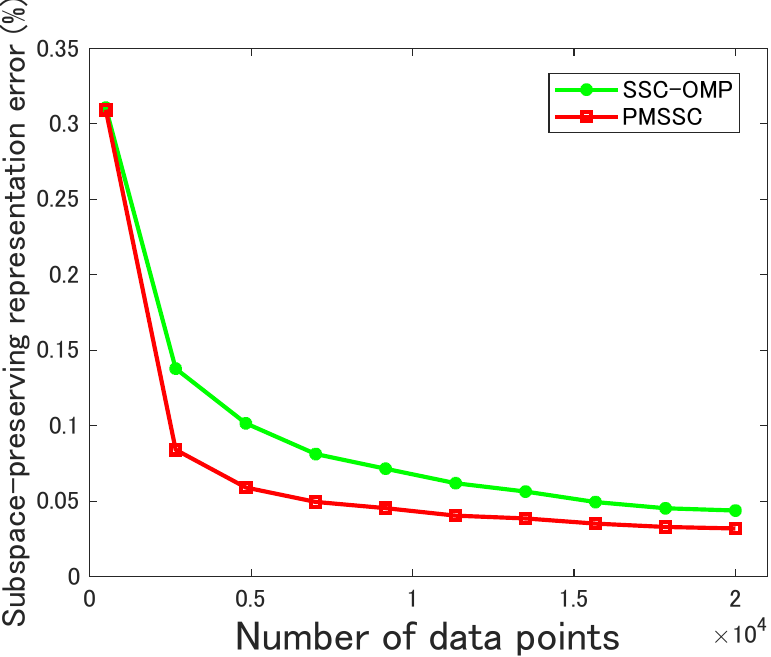}
    \label{fig:res_syn_b}
    }\\
    \subfigure[]{\includegraphics[width=0.355\linewidth]{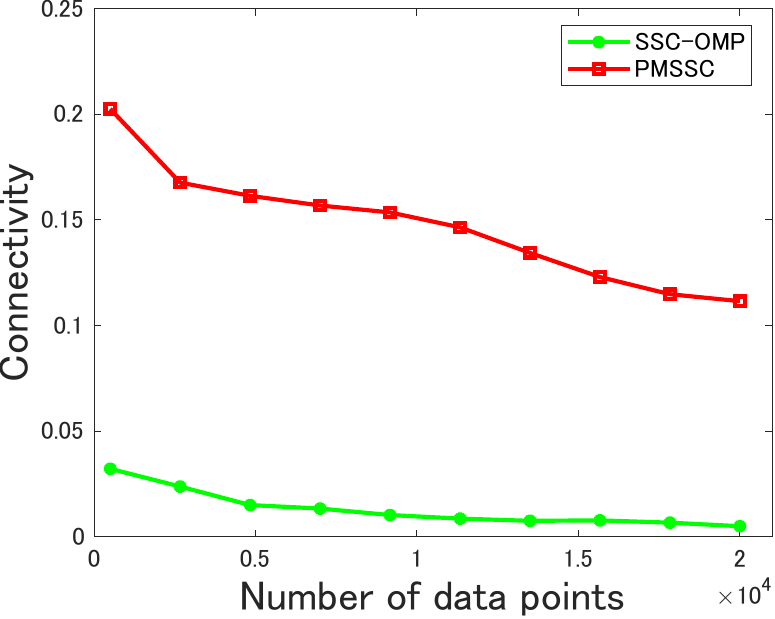}
    \label{fig:res_syn_c}
    }\hspace{10pt}
    \subfigure[]{\includegraphics[width=0.35\linewidth]{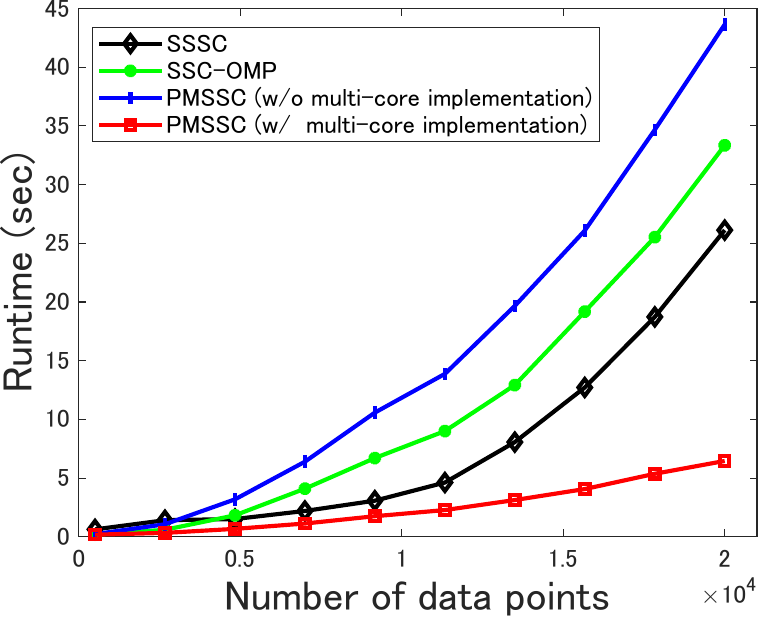}
    \label{fig:res_syn_d}
    }
    \caption{Comparison of PMSSC, SSC-OMP, and SSSC on synthetic data in terms of (a) clustering accuracy, (b) subspace-preserving representation error, (c) connectivity, and (d) runtime. For SSSC, only  clustering accuracy and runtime are shown as SSSC does not generate the self-expressive coefficient matrix and affinity matrix.
    }
    \label{fig:res_syn}
\end{figure*}

\begin{figure*}[t]
    \centering
    \subfigure[]{\includegraphics[width=0.22\linewidth]{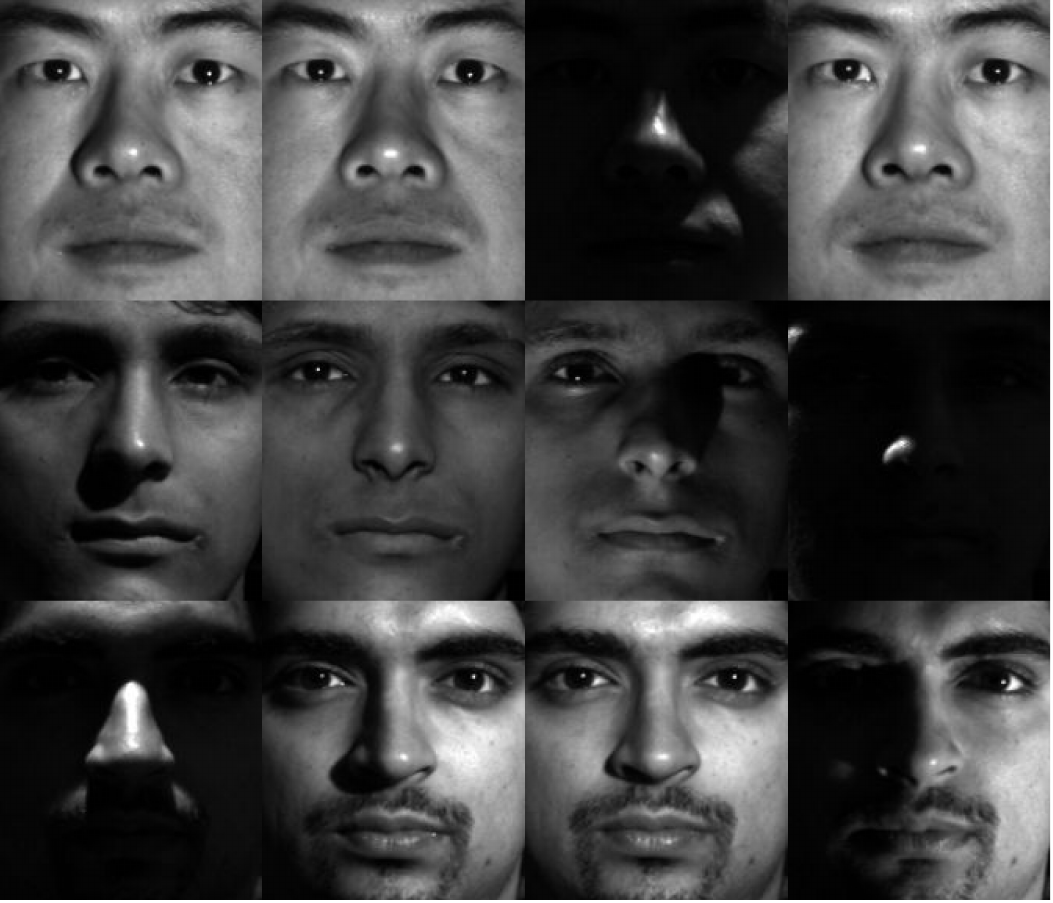}
    \label{fig:Datasets_EYaleB}
    }
    \subfigure[]{\includegraphics[width=0.26\linewidth]{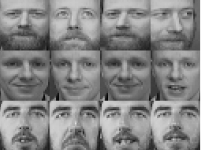}
    \label{fig:Datasets_ORL}
    }
    \subfigure[]{\includegraphics[width=0.26\linewidth]{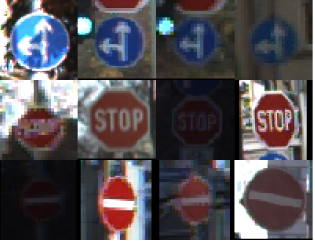}
    \label{fig:Datasets_GTSRB}
    }
    \subfigure[]{\includegraphics[width=0.26\linewidth]{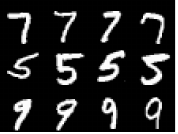}
    \label{fig:Datasets_MNIST}
    }
    \subfigure[]{\includegraphics[width=0.26\linewidth]{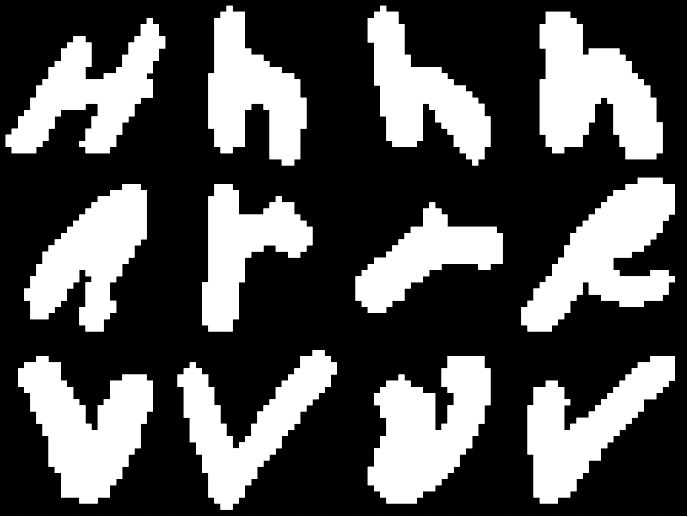}
    \label{fig:Datasets_EMNIST}
    }
    \subfigure[]{\includegraphics[width=0.26\linewidth]{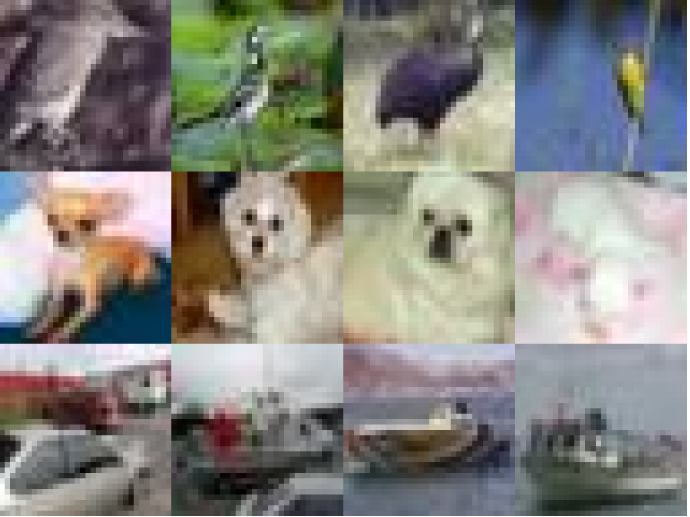}
    \label{fig:Datasets_CIFAR-10}
    }
    \caption{Visual examples for datasets: (a) Extended Yale B, (b) ORL, (c) GTSRB, (d) MNIST, (e) EMNIST-Letters, and (f) CIFAR-10.}
    \label{fig:Datasets}
\end{figure*}

\begin{table*}[t!]
\centering
\caption{Parameters ($s$, $\delta$, and $T$) used in PMSSC for benchmark datasets.}
\label{tab:OurParam}
\begin{tabular}{c|ccccccccc}
\hline
Datasets & Ex. Yale B & ORL & GTSRB & BBCSport & MNIST4000 & MNIST10000 & MNIST & EMNIST & CIFAR-10 \\ \hline
$s$      & 5   & 5   & 3   & 3   & 10  & 10  & 10 & 10 & 3\\
$\delta$ & 0.6 & 0.6 & 0.2 & 0.4 & 0.3 & 0.2 & 0.1 & 0.2 & 0.2\\
$T$      & 6   & 11  & 8  & 15  & 7  & 10  & 19 & 12 &  18\\ \hline
\end{tabular}
\end{table*}

\subsection{Experiments on Synthetic Data}
\label{sec:sec4_1}
\subsubsection{Setup}
We first report experimental results on data synthesised by randomly generating five linear subspaces of $\mathbb{R}^{6}$ as ground-truth in an ambient space of $\mathbb{R}^{9}$.
Each subspace contains $n$ randomly sampled data points.
To confirm the statistical results, we conducted the experiments by varying $n$ from 100 to 4,000, so the total number $N$ of data points varies from 500 to 20,000.
We set the parameters $s = 6$, $\delta = 0.3$, and $T = 16$.
The percentage of in-sample in SSSC is set to $10\%$ of the total number of data points.
All experimental results recorded on synthetic data were averaged over 50 trials.

\subsubsection{Results}
The curve for each metric is shown as a function of $n$ in Fig.~\ref{fig:res_syn}.
We can observe from Fig.~\ref{fig:res_syn_a} that PMSSC outperforms SSC-OMP in terms of clustering accuracy.
The difference is especially large when the density of data points on the underlying subspace is lower.
This could be partly due to the fact that PMSSC succeeds in generating better connectivity than SSC-OMP (see Fig.~\ref{fig:res_syn_c}), and achieves lower subspace-preserving representation error (see Fig.~\ref{fig:res_syn_b}).
On the other hand, as Fig.~\ref{fig:res_syn_d} shows, PMSSC is much faster with parallel implementation, which is advisable for solving problems involving large-scale data.
In addition, compared to SSSC which adopts a similar sampling approach to our method, PMSSC outperforms both in clustering accuracy and runtime (using a parallel implementation).

\subsection{Experiments on Benchmark Datasets for Real-world Applications}
\label{sec:sec4_2}
\subsubsection{Setup}
We conducted experiments on five benchmark datasets: Extended Yale B~\cite{lee2005acquiring} and ORL~\cite{samaria1994parameterisation} for face clustering, BBCSport~\cite{greene06icml} for text document clustering, German Traffic Sign Recognition Benchmark (GTSRB)~\cite{stallkamp2012man} for street sign clustering, Modified National Institute of Standards and Technology database (MNIST)~\cite{lecun1998gradient} and Extended MNIST (EMNIST)~\cite{cohen2017emnist} for handwritten character clustering, and Canadian Institute For Advanced Research (CIFAR-10)~\cite{krizhevsky2009learning} for object clustering.
Parameter settings used for our method in these experiments are shown in Table~\ref{tab:OurParam}.
Since the sparsity $s$ in PMSSC and SSC-OMP is related to the intrinsic dimensionality of the subspace, it is manually determined to be close to the dimensionality of the subspaces.
For sampling rate $\delta$, we picked a smaller $\delta$ for a larger dataset.
For the number of subsets $T$, we adopted a value that takes into account the trade-off between runtime and clustering accuracy.
All experimental results recorded on all benchmark datasets are averaged over 10 trials.
Details of each benchmark dataset and the corresponding clustering results are now given.

\begin{table}[t]
\centering
\caption{Comparative results on Extended Yale B; `-' indicates the metric cannot be computed.}
\label{tab:EYaleB}
\begin{tabular}{l|rrrr}
\hline
\multirow{2}{*}{Method} & \multicolumn{4}{c}{Extended Yale B} \\
                        & acc ($a\%$) & sre ($e\%$)       & conn ($c$)& $t$ (sec.) \\ \hline
SCC                     & 9.54  & -     & -    & 293.01 \\
LRR                     & 37.58 & 97.44 & \textbf{0.8175} & 219.91 \\
TSC                     & 52.40 & -     & 0.0014    & 10.46  \\
LRSC                    & 56.71 & 91.64 & \underline{0.4360} & 4.37 \\
SSSC                    & 49.77 & -     & -    & 18.22   \\
EnSC                    & 55.76 & \textbf{18.90} & 0.0395 & 4.57   \\
SSC-OMP                 & \underline{73.82} & \underline{20.07} & 0.0364 & \textbf{1.27}   \\
S$^5$C                  & 62.99 & 58.74 & 0.2238 & 952.26 \\ \hline
PMSSC                    & \textbf{80.24} & 22.35 & 0.0858 & \underline{2.66} \\ \hline
\end{tabular}
\end{table}
\begin{table}[t]
\centering
\caption{Comparative results on ORL.}
\label{tab:ORL}
\begin{tabular}{l|rrrr}
\hline
\multirow{2}{*}{Method} & \multicolumn{4}{c}{ORL} \\
                        & acc ($a\%$) & sre ($e\%$)       & conn ($c$)& $t$ (sec.) \\ \hline
SCC                     & 16.40 & -     & -      & 89.45 \\
LRR                     & 32.98 & 97.70 & \textbf{0.8394} & 3.57 \\
TSC                     & 68.03 & -     & 0.0992      & 0.72  \\
LRSC                    & 43.12 & 93.72 & \underline{0.5248} & \underline{0.24} \\
SSSC                    & 65.12 & -     & -      & 1.78   \\
EnSC                    & \underline{70.03} & \textbf{32.46} & 0.1825 & 0.65   \\
SSC-OMP                 & 60.12 & \underline{34.14} & 0.0770 & \textbf{0.12}   \\
S$^5$C                  & 69.48 & 63.26 & 0.3868 & 54.28 \\ \hline
PMSSC                    & \textbf{74.45} & 40.97 & 0.1708 & 0.51 \\ \hline
\end{tabular}
\end{table}

\begin{table}[t]
\centering
\caption{Comparative results on GTSRB.}
\label{tab:GTSRB}
\begin{tabular}{l|rrrr}
\hline
\multirow{2}{*}{Method} & \multicolumn{4}{c}{GTSRB} \\
                        & acc ($a\%$) & sre ($e\%$)       & conn ($c$)& $t$ (sec.) \\ \hline
SCC                     & 59.68 & -     & -    & 84.26 \\
LRR                     & 27.87 & 86.64 & 0.4255 & 725.18 \\
TSC                     & 56.36 & -     & 0.0016 & 242.62  \\
LRSC                    & 83.97 & 80.14 & \textbf{0.6056} & 12.89 \\
SSSC                    & \underline{88.03} & -     & -      & 16.86   \\
EnSC                    & 85.92 & \textbf{0.59}  & 0.0065 & 10.77   \\
SSC-OMP                 & 81.28 & \underline{5.38}  & 0.0211 & \underline{3.72}   \\
S$^5$C                  & 61.60 & 80.99 & \underline{0.5941} & 422.35 \\ \hline
PMSSC                    & \textbf{91.57} & 7.69 & 0.0434  & \textbf{3.40} \\ \hline
\end{tabular}
\end{table}
\begin{table}[t]
\centering
\caption{Comparative results on BBCSport.}
\label{tab:BBCSport}
\begin{tabular}{l|rrrr}
\hline
\multirow{2}{*}{Method} & \multicolumn{4}{c}{BBCSport} \\
                        & acc ($a\%$) & sre ($e\%$)       & conn ($c$)& $t$ (sec.) \\ \hline
SCC                     & 23.12 & -     & -      & 3.60 \\
LRR                     & 71.37 & 76.26 & \textbf{0.7744} & 7.59 \\
TSC                     & 73.95 & -     & 0.0053 & 0.36  \\
LRSC                    & \textbf{89.53} & 66.38 & \underline{0.5997} & \underline{0.18} \\
SSSC                    & 50.24 & -     & -      & 0.26   \\
EnSC                    & 59.48 & \textbf{11.43} & 0.0243 & 0.61   \\
SSC-OMP                 & 69.85 & 15.96 & 0.0393 & \textbf{0.10}   \\
S$^5$C                  & 55.90 & 65.78 & 0.5434 & 17.99 \\ \hline
PMSSC                    & \underline{81.71} & \underline{14.36} & 0.0509 & 0.47 \\ \hline
\end{tabular}
\end{table}

\begin{table*}[t]
\centering
\caption{Comparative results on MNIST4000 and MNIST10000.}
\label{tab:MNIST1}
\begin{tabular}{l|rrrr|rrrr}
\hline
\multirow{2}{*}{Method} & \multicolumn{4}{c|}{MNIST4000} & \multicolumn{4}{c}{MNIST10000}                                            \\
                        & acc ($a\%$) & sre ($e\%$) & conn ($c$)& $t$ (sec.) & acc ($a\%$) & sre ($e\%$) & conn ($c$) & $t$ (sec.) \\ \hline
SCC                     & 67.45 & -     & -      & 5.93   & 70.43 & - & - & 11.67 \\
LRR                     & 78.49 & 90.21 & \textbf{0.8979} & 43.03  & 77.53 & 90.60 & \textbf{0.8818} & 396.45 \\
TSC                     & 79.57 & -     & 0.0009 & 11.76  & 80.62 & - & 0.0005 & 132.08 \\
LRSC                    & 81.23 & 75.67 & \underline{0.5984} & \underline{1.61}   & 80.86 & 77.30 & \underline{0.5983} & 7.58 \\
SSSC                    & 70.73 & -     & -      & 3.11   & 84.32 & -     & -    & 13.20 \\
EnSC                    & 89.08 & \textbf{21.14} & 0.1174 & 12.71  & 88.24 & \textbf{17.34}     & 0.0975    & 35.63 \\
SSC-OMP                 & \underline{91.49} & \underline{34.69} & 0.1329 & \underline{1.61}   & \underline{91.40} & \underline{32.23}     & 0.1169    & \underline{6.44} \\
S$^5$C                  & 81.52 & 66.28 & 0.4476 & 277.93 & 79.30 & 66.23     & 0.4466    & 683.65 \\ \hline
PMSSC                    & \textbf{92.85} & 38.27 & 0.1944 & \textbf{1.42}   & \textbf{93.57} & 36.43     & 0.1817    & \textbf{4.55} \\ \hline
\end{tabular}
\end{table*}

\begin{table}[t]
\centering
\caption{Comparative results on MNIST; `M' indicates that 32 GB memory was exhausted.}
\label{tab:MNIST2}
\begin{tabular}{l|rrrr}
\hline
\multirow{2}{*}{Method} & \multicolumn{4}{c}{MNIST70000}                                           \\
\multicolumn{1}{c|}{}   & acc ($a\%$) & sre ($e\%$)       & conn ($c$)& $t$ (sec.) \\ \hline
SCC                     & 69.08 & -     & -    & 388.00 \\
LRR                     & M & - & - & - \\
TSC                     & M & -     & -    & -  \\
LRSC                    & M & - & - & - \\
SSSC                    & 81.57 & -     & -    & 303.28   \\
EnSC                    & \textbf{93.79} & \textbf{11.26} & 0.0596 & 408.62   \\
SSC-OMP                 & 82.83 & \underline{28.57} & 0.0830 & \underline{248.50}   \\
S$^5$C                  & 72.99 & 66.87 & \textbf{0.4437} & 4953.28 \\ \hline
PMSSC                    & \underline{84.45} & 32.63 & \underline{0.1148} & \textbf{65.08} \\ \hline
\end{tabular}
\end{table}

\subsubsection{Extended Yale B}
Extended Yale B contains 2,432 facial images in 38 classes; see Fig.~\ref{fig:Datasets_EYaleB}. In this experiment, following~\cite{elhamifar2013sparse}, we concatenated the pixels of each image resized to $48 \times 42$, and used the 2016-dimensional vector as input data.

The results on Extended Yale B are shown in Table~\ref{tab:EYaleB}.
In each column, the best result is shown in bold, and the second-best result is underlined.
They confirm that PMSSC yields the best clustering accuracy, and improves the clustering accuracy over SSC-OMP by $6.42\%$.
Although the subspace-preserving error and runtime are slightly lower than SSC-OMP, the connectivity is greatly improved compared to SSC-OMP, leading to a better clustering accuracy.
LRR, LRSC, and S$^5$C have good connectivity, but poor subspace-preserving errors result in low clustering accuracy.

\subsubsection{ORL}
ORL contains 400 facial images in 40 classes, as shown in Fig.~\ref{fig:Datasets_ORL}.
In this experiment, following~\cite{cai2007learning}, we concatenate the pixels of each image resized to $32 \times 32$, and use a 1024-dimensional vector as input data.
Compared to Extended Yale B, ORL is a more difficult problem setting for subspace clustering because the density of data lying near the same subspace (10 vs. 64) is lower due to the small number of images of each subject, and the subspaces have more non-linearity due to changes in facial expressions and details.

The results for ORL are listed in Table~\ref{tab:ORL}.
We can again observe that PMSSC achieves the best clustering accuracy, and improves the connectivity compared to SSC-OMP.
However, since PMSSC does not incorporate nonlinear constraints, the subspace-preserving error does not improve along with the improvement of the connectivity.

\subsubsection{GTSRB}
GTSRB contains over 50,000 street sign images in 43 classes; see Fig.~\ref{fig:Datasets_GTSRB}.
Following~\cite{you2018scalable}, we preprocess the dataset represented by a 1568-dimensional HOG feature to get an imbalanced dataset of the 500-dimensional vectors with 12,390 samples in 14 classes.

The results on GTSRB are reported in Table~\ref{tab:GTSRB}.
Again PMSSC yields the best clustering accuracy and runtime, and improves the clustering accuracy roughly by $10\%$ compared to SSC-OMP.
In particular, PMSSC has both good subspace-preserving error and connectivity.
While EnSC and SSSC also achieve competitive clustering accuracy, their computational costs are much higher.

\begin{table}[t]
\centering
\caption{Comparative results on EMNIST-Letters; `M' indicates that 32 GB memory was exhausted.}
\label{tab:EMNIST}
\begin{tabular}{l|rrrr}
\hline
\multirow{2}{*}{Method} & \multicolumn{4}{c}{EMNIST-Letters}                                           \\
\multicolumn{1}{c|}{}   & acc ($a\%$) & sre ($e\%$)       & conn ($c$)& $t$ (sec.) \\ \hline
SCC                     & M & - & - & - \\
LRR                     & M & - & - & - \\
TSC                     & M & - & - & - \\
LRSC                    & M & - & - & - \\
SSSC                    & 60.62 & -     & -    & 1538.46   \\
EnSC                    & \underline{64.15} & \textbf{26.20} & \underline{0.0086} & 1575.46 \\
SSC-OMP                 & 58.71 & \underline{43.93} & 0.0000 & \underline{1214.31}   \\
S$^5$C                  & 60.01 & 83.37 & \textbf{0.3517} & 15698.90 \\ \hline
PMSSC                   & \textbf{66.52} & 46.76 & 0.0019 & \textbf{638.03} \\ \hline
\end{tabular}
\end{table}

\subsubsection{BBCSport}
BBCSport contains 737 documents in five classes.
The data provided by the database has been preprocessed by stemming, stop-word removal, and low term frequency filtering.
In this experiment, we reduced the dimensionality of feature vectors to 500 by PCA.

The results on BBCSport are summarized in Table~\ref{tab:BBCSport}.
We can observe that PMSSC yields the second best clustering accuracy and subspace-preserving error.
LRSC yields the best clustering accuracy due to good connectivity.
For small-scale datasets such as BBCSport and ORL, the speed of PMSSC is slightly lower than for SSC-OMP because the advantage of reducing data size by sampling multiple subsets is diminished.

\subsubsection{MNIST and EMNIST-Letters}
MNIST contains 70,000 images of handwritten digits (0--9), while EMNIST-Letters contains 145,600 images of handwritten characters in 26 classes, as shown in Figs.~\ref{fig:Datasets_MNIST} and \ref{fig:Datasets_EMNIST}.
In our experiments, following~\cite{chen2020stochastic}, we generate MNIST4000 and MNIST10000, which are produced by randomly sampling 400 and 1,000 images per class of digit, respectively.
Each image is represented as a 3472-dimensional feature vector by using the scattering convolution network~\cite{bruna2013invariant}, and its dimensionality reduced to 500 by PCA.

The results on MNIST and EMNIST-Letters are summarized in Tables~\ref{tab:MNIST1}--\ref{tab:EMNIST}.
We can observe that PMSSC yields the best clustering accuracy on MNIST4000, MNIST10000, and EMNIST-Letters.
In particular,  PMSSC is remarkably faster than the comparative methods on MNIST70000 and EMNIST-Letters.
In the case of MNIST70000, EnSC yields the best clustering accuracy and subspace-preserving error but its computational cost is high.
Similarly, S$^5$C can achieve good connectivity, but is very slow.

\subsubsection{CIFAR-10}
CIFAR-10 includes 60,000 general objects in 10 classes, as illustrated in Fig.~\ref{fig:Datasets_CIFAR-10}.
Following~\cite{zhang2021learning}, we employ the feature representations extracted by MCR$^{2}$~\cite{yu2020learning}, which learns a union of low-dimensional subspaces representation  via self-supervised learning.
Each feature is represented by a 128-dimensional feature vector, further normalized to have unit $\ell_{2}$ norm.

The comparative results on CIFAR-10 are summarized in Table~\ref{tab:CIFAR-10}.
It can be observed that our method outperforms others in terms of the runtime, while the clustering accuracy is competitive.
However, as with SSC-OMP, we see that the connectivity is lower than for S$^{5}$C, which uses $\ell_{1}$ norm regularization.

\subsubsection{Summary}
Overall, our proposed method becomes significantly faster as the amount of input data increases. In addition, it achieves good clustering accuracy and connectivity, and provides subspace-preserving errors comparable to those of the comparative algorithms.

\begin{table}[t]
\centering
\caption{Comparative results on CIFAR-10 where 'M' means that the memory limitation of 32G is exceeded.}
\label{tab:CIFAR-10}
\begin{tabular}{l|cccc}
\hline
\multirow{2}{*}{Method} & \multicolumn{4}{c}{CIFAR-10}                                           \\
\multicolumn{1}{c|}{}   & acc ($a\%$) & sre ($e\%$)       & conn ($c$)& $t$ (sec.) \\ \hline
SCC                     & 37.10 & -     & -    & 196.40 \\
LRR                     & M & - & - & - \\
TSC                     & M & -     & -    & -  \\
LRSC                    & M & - & - & - \\
SSSC                    & \underline{63.80} & -     & -    & 74.36   \\
EnSC                    & 61.79 & \textbf{22.60} & 0.0000 & 178.22 \\
SSC-OMP                 & 40.86 & \underline{24.92} & 0.0000 & \underline{63.58}   \\
S$^5$C                  & \textbf{64.52} & 46.35 & \textbf{0.2314} & 2338.55 \\ \hline
PMSSC                   & 63.52 & 26.41 & 0.0000 & \textbf{29.60} \\ \hline
\end{tabular}
\end{table}

\subsection{Analysis}
\label{sec:sec4_3}
\subsubsection{Multi-subset Based Self-Expressive Model}
\begin{figure}[t]
\centering
    \includegraphics[width=\linewidth]{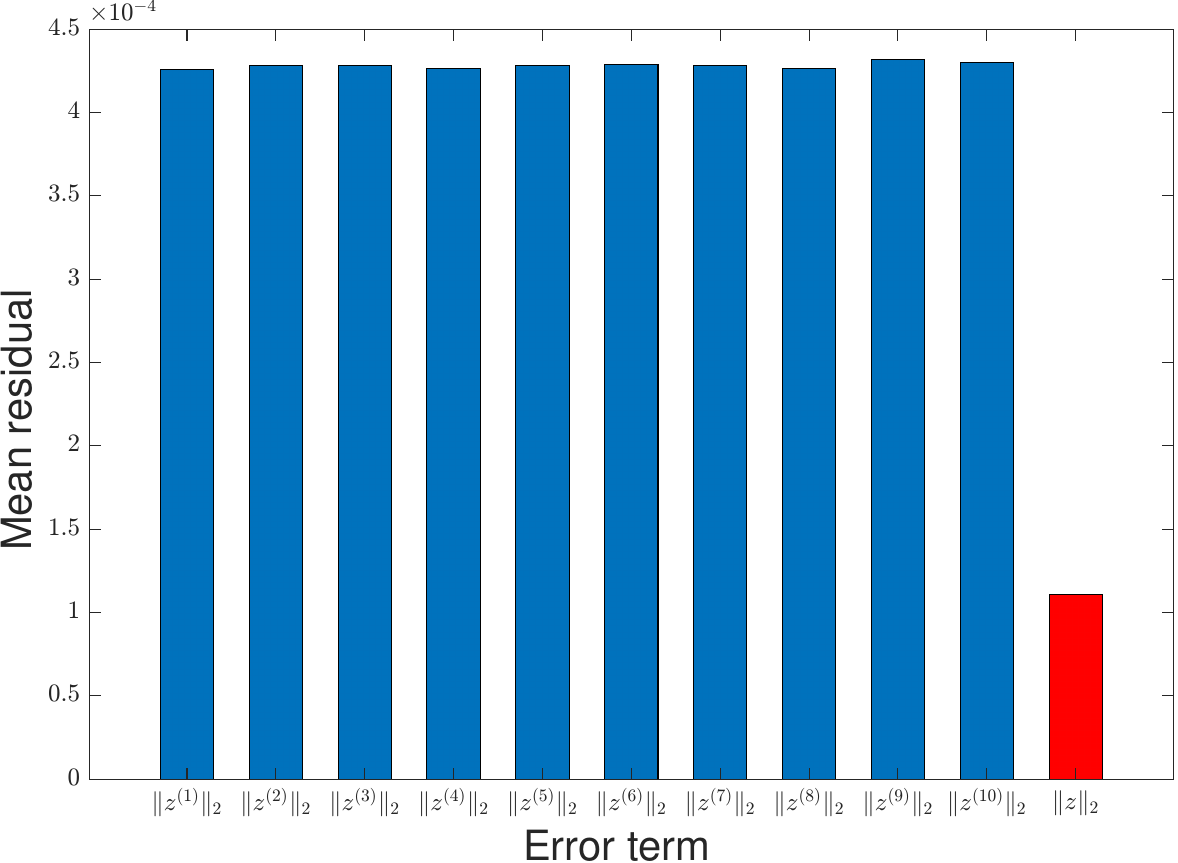}
  \caption{Comparative results in terms of the mean residuals over data points represented by the self-expressive models with different subsets. Blue bars represent each single $t$-th subset, while the red bar is computed using multiple subsets.}
  \label{fig:Analysis2}
\end{figure}

\begin{figure}[t]
\centering
    \subfigure[]{\includegraphics[width=0.47\linewidth]{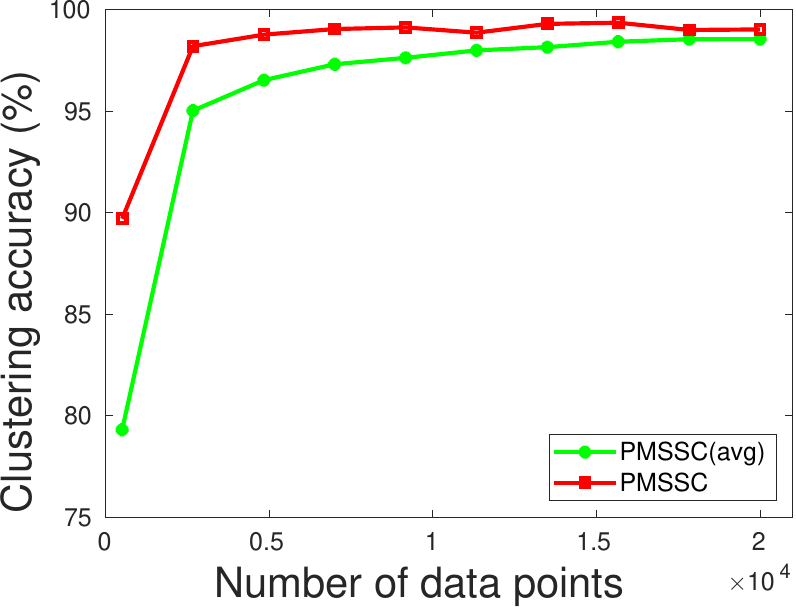}
    \label{fig:ours_vs_rs_accuracy}
    }
    \subfigure[]{\includegraphics[width=0.47\linewidth]{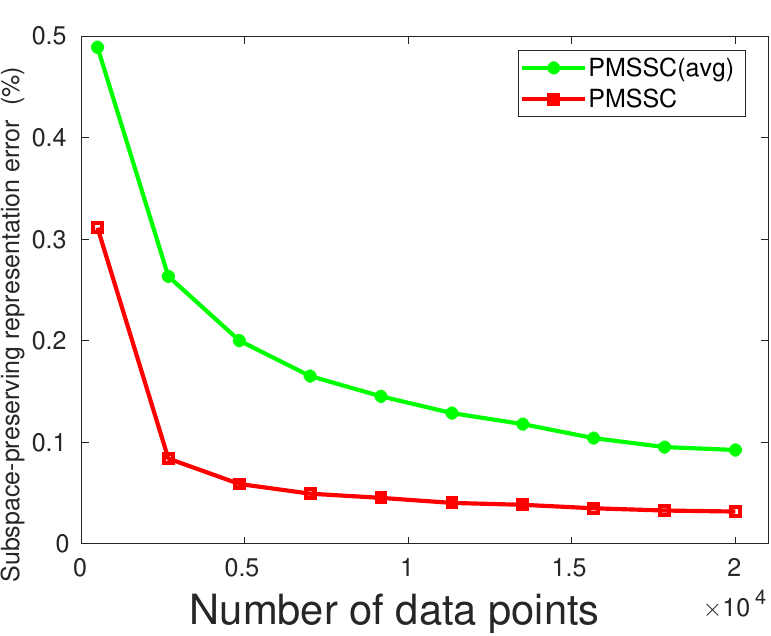}
    \label{fig:ours_vs_rs_sre}
    }
  \caption{Benefit of using $\boldsymbol{b}^{\ast}$ in PMSSC in terms of (a) clustering accuracy and (b) subspace-preserving representation error, for synthetic data. Red: using $\boldsymbol{b}^{\ast}$. Green: using simple averaging. }
  \label{fig:Ablation}
\end{figure}

Since our approach aims to minimize the self-expressive residual by the weight coefficient vector $\boldsymbol{b}^{\ast}$ solved in Algorithm~\ref{alg:alg3}, we show the mean self-expressive residual of data points represented by the coefficient vectors in Fig.~\ref{fig:Analysis2}.
This experiment was performed on synthetic data, and we fixed $T=10$ and $\delta=0.3$.
Each blue bar indicates the mean self-expressive residual of the data points represented by Eq. (\ref{equ:equ5}),  computed as
\begin{equation}
    \begin{split}
    \|\boldsymbol{z}^{(t)}\|_{2} = \frac{1}{\lceil \delta N \rceil}\sum_{i=1}^{N} \|\boldsymbol{x}_{i}^{(t)} - X^{(t)} \boldsymbol{c}_{i}^{\ast(t)}\|_{2}.
    \end{split}
    \label{equ:residual_sub}
\end{equation}
The red bar indicates the mean self-expressive residual of the data points represented by Eq. (\ref{equ:equ9}),  computed as: 
\begin{equation}
    \|\boldsymbol{z}\|_{2} = \frac{1}{N}\sum_{i=1}^{N}\|(\boldsymbol{x}_{i} - X \boldsymbol{c}_{i}^{\ast})\|_{2}.
    \label{equ:residual_ours}
\end{equation}
We can clearly observe that the mean self-expressive residual of PMS has lower error than every single subset. 
To highlight the benefit of $\boldsymbol{b}^{\ast}$, we made a comparison to a variant of our approach, named PMSSC(avg), which replaced $\boldsymbol{b}^{\ast}$ by a simple average operation:
in PMSSC(avg), Eq. (\ref{equ:equ10}) is replaced by
\begin{equation}
    \boldsymbol{c}_{i}^{\ast} = \frac{1}{T} \sum_{t=1}^{T} \boldsymbol{c}_{i}^{\ast (t)}.
    \label{equ:pmssc_avg}
\end{equation}
We performed experiments on synthetic data using the same setup as for Fig.~\ref{fig:res_syn} and present the results in Fig.~\ref{fig:Ablation}.
As can be seen, incorporating $\boldsymbol{b}^{\ast}$ improves  clustering performance; in particular, the subspace-preserving representation error is significantly reduced.
These experiments indicate that the weight coefficient vector $\boldsymbol{b}^{\ast}$ contributes to improving self-expressiveness.

\begin{figure}[t]
    \centering
    \subfigure[]{\includegraphics[width=0.47\linewidth]{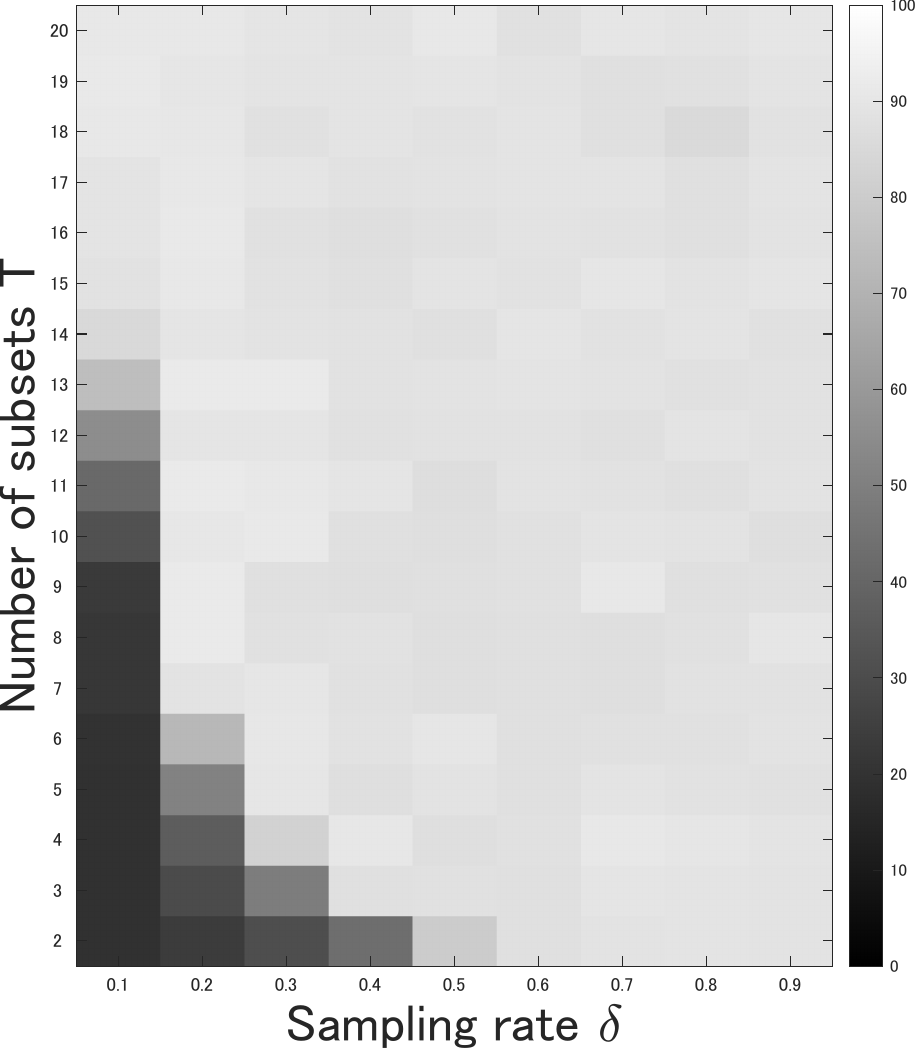}
    \label{fig:accuracy_GTSRB}
    }
    \subfigure[]{\includegraphics[width=0.47\linewidth]{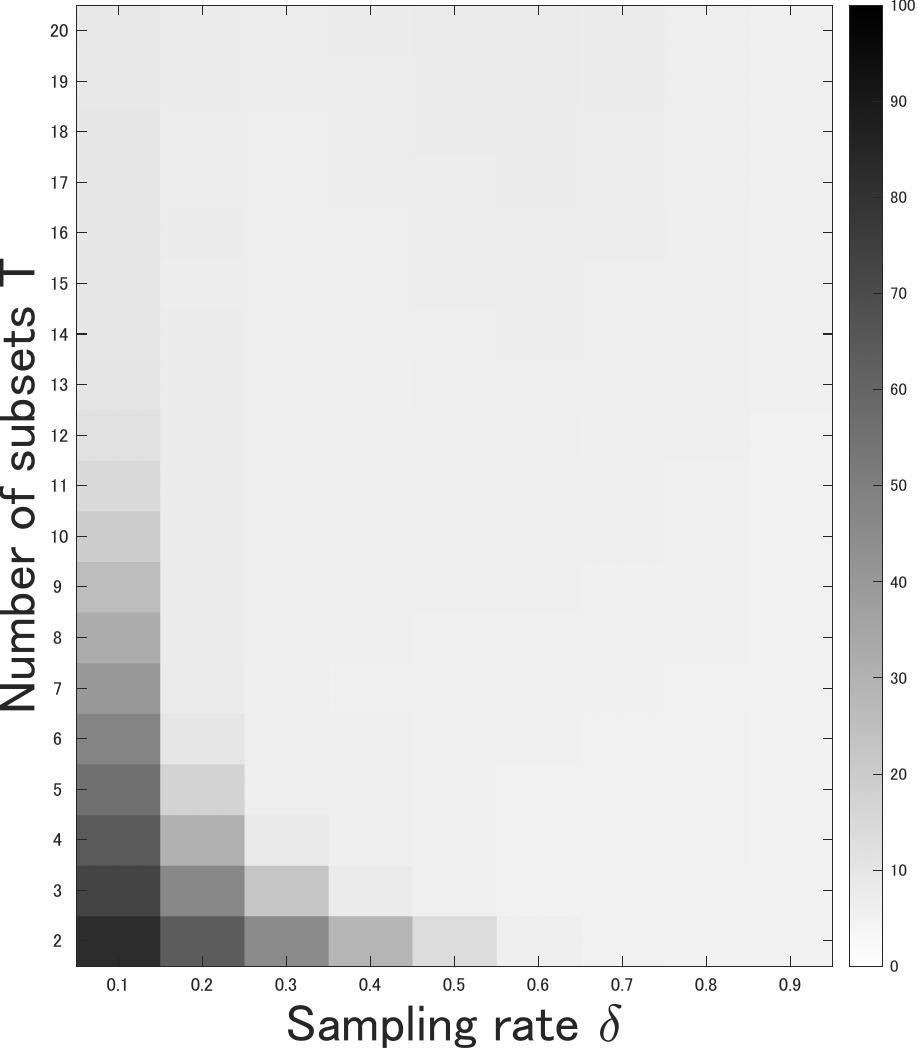}
    \label{fig:sre_GTSRB}
    }
    \subfigure[]{\includegraphics[width=0.47\linewidth]{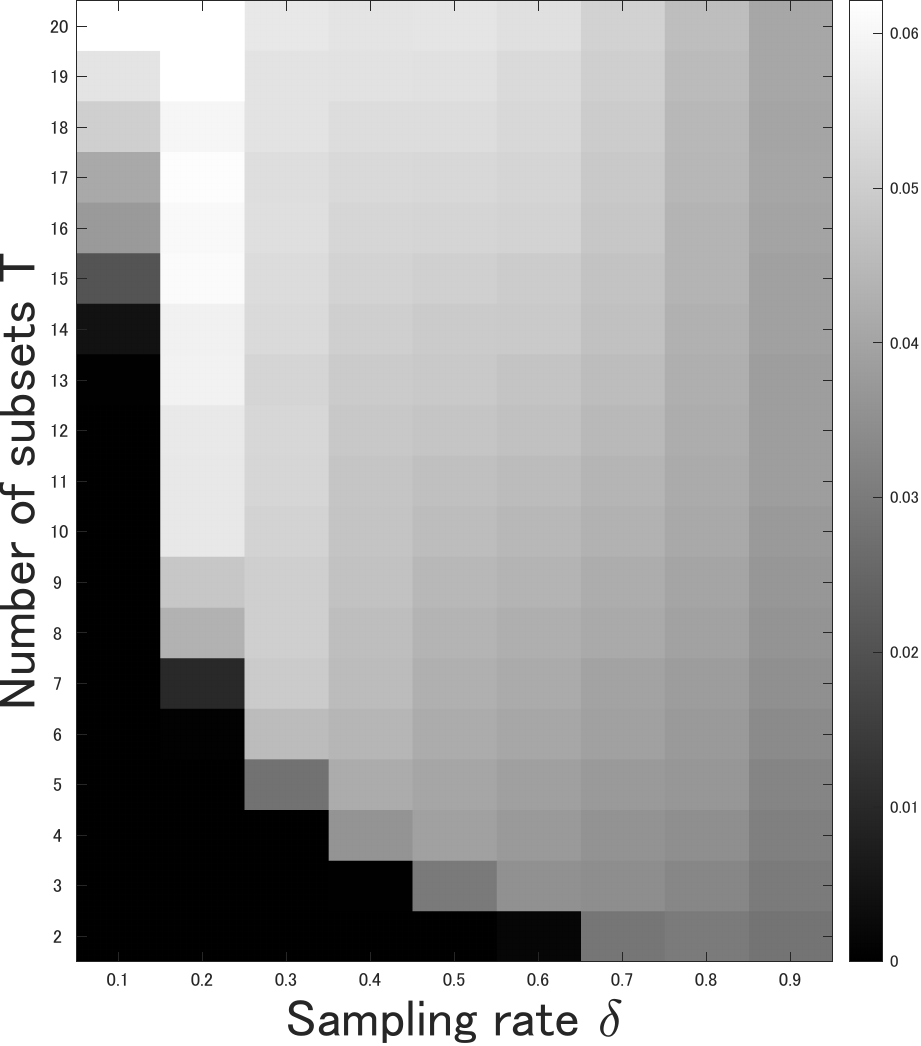}
    \label{fig:connectivity_GTSRB}
    }
    \subfigure[]{\includegraphics[width=0.47\linewidth]{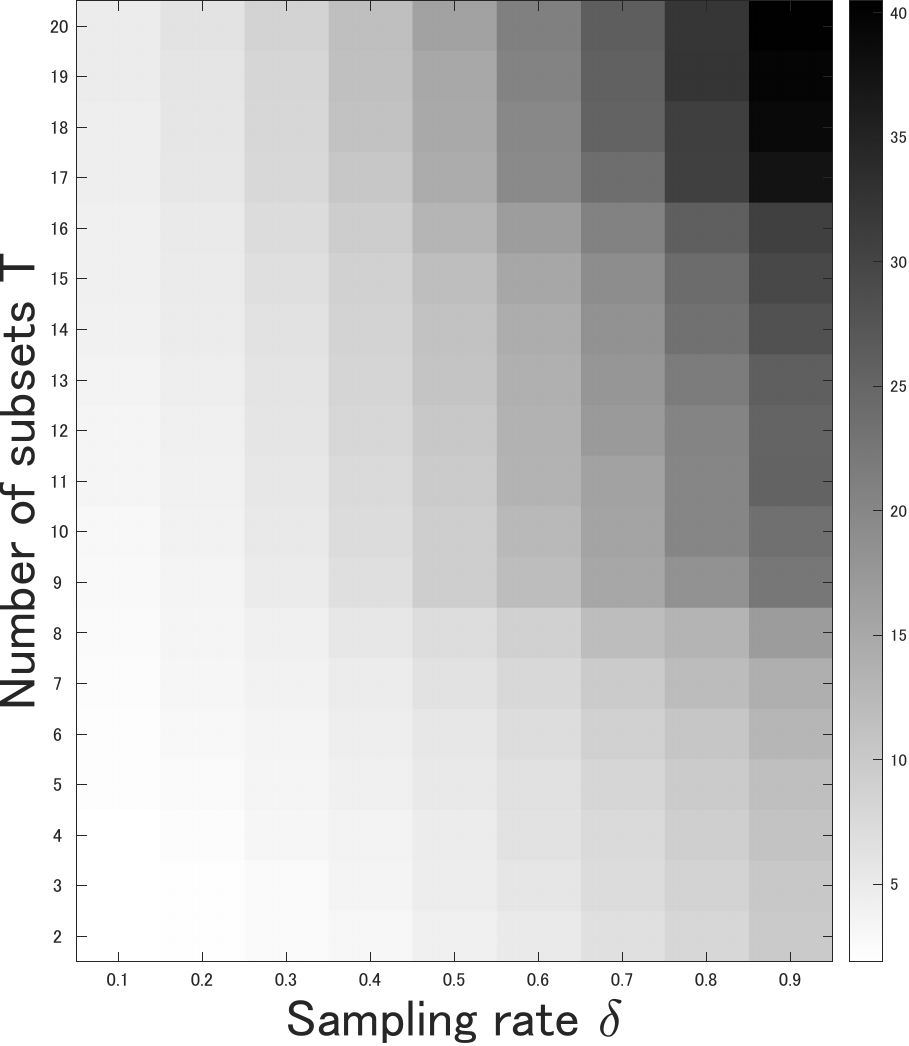}
    \label{fig:running_time_GTSRB}
    }
    \caption{Effects of varying parameters $\delta$ and $T$ (GTSRB dataset): (a) clustering accuracy, (b) subspace-preserving representation error, (c) connectivity, and (d) runtime.}
    \label{fig:Analysis_GTSRB}
\end{figure}

\begin{figure}[t]
    \centering
    \subfigure[]{\includegraphics[width=0.48\linewidth]{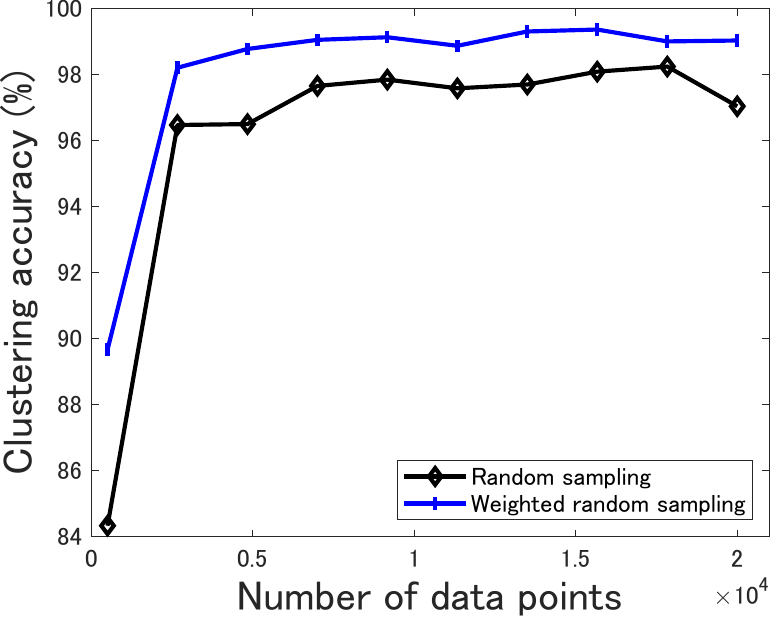}
    \label{fig:accuracy}
    }
    \subfigure[]{\includegraphics[width=0.48\linewidth]{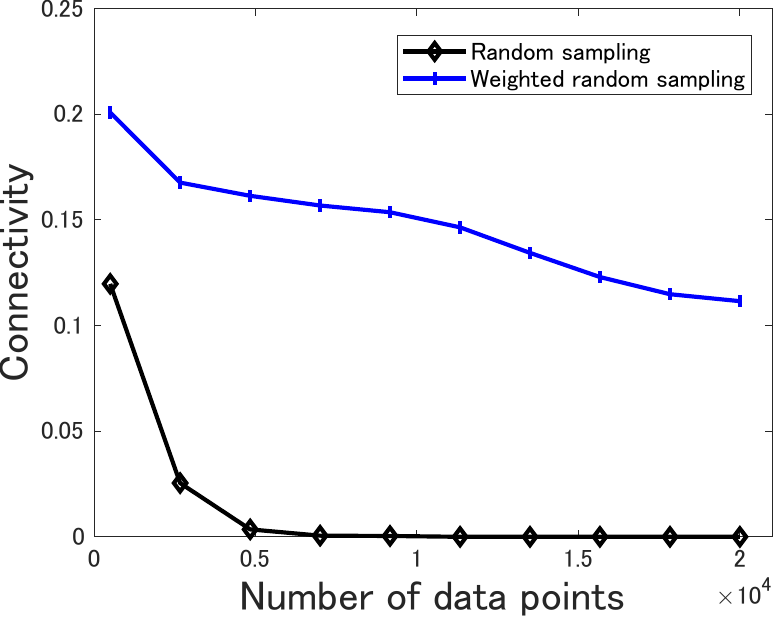}
    \label{fig:running_time}
    }
    \caption{Effect of sampling method in our approach, for synthetic data: (a) clustering accuracy and (b) connectivity. Blue: weighted random sampling. Black: uniform random smapling.}
    \label{fig:Analysis_sampling}
\end{figure}

\subsubsection{Selection of Parameters}
We performed multiple experiments on the GTSRB dataset with various choices of hyperparameters $(T,\delta)$ to evaluate the sensitivity of our approach to parameter choice. 
Changes in clustering accuracy, subspace-preserving representation error, connectivity, and runtime when varying each parameter are illustrated in Fig.~\ref{fig:Analysis_GTSRB}.
We can confirm that high clustering accuracy and low subspace-preserving representation error are maintained in most cases, except when both $T$ and $\delta$ are extremely small.
This implies that the affinity matrix constructed by PMSSC provides subspace-preserving representations at most data points.
We can also see that the connectivity improves as the number of subsets $T$ increases, because the affinity matrix contains at most $sTN$ nonzero entries in OMP optimization.
Considering  runtime, a practical choice of parameters is to increase $T$ for small values of $\delta$, and decrease $T$ for large values of $\delta$. In addition, time taken can be kept low by picking a small value of $\delta$ for large-scale datasets.

\subsubsection{Sampling Technique}
Our approach adopts  weighted random sampling to generate the subset data matrix $X^{(t)}$. 
To analyze the effect of sampling methods on our approach, we compared  weighted random sampling to random sampling with uniform weights.
The experimental settings used for synthetic data follow those in Fig.~\ref{fig:res_syn}. 
Fig.~\ref{fig:Analysis_sampling} shows the clustering accuracy and  connectivity as functions of $n$.
Obviously, weighted random sampling outperforms random sampling in terms of both  clustering accuracy and  connectivity. In particular, as the density of data points increases, the connectivity of the method with random sampling becomes zero, because  imbalanced sampling leads to a disconnected subgraph in an affinity graph.

\subsubsection{Computational Complexity}
Algorithms~\ref{alg:alg2} and~\ref{alg:alg3} for affinity matrix construction consume most of the processing time.
In Algorithm~\ref{alg:alg2}, the computational complexity for finding the self-expressive coefficient vector $\boldsymbol{c}_{i}^{\ast(t)}$ requires time $\mathcal{O}(Ds\lceil \delta N \rceil)$.
In Algorithm~\ref{alg:alg3}, the computational complexity for finding the weight coefficient vector $\boldsymbol{b}^{\ast(i)}$ requires $\mathcal{O}(DT^{2})$.
Because these two algorithms are performed on $N$ data points, the computational complexity of PMS requires at least time $\mathcal{O}(N(TDs\lceil \delta N \rceil + DT^{2}))$. However, processing $T$ subsets (the part taking $\mathcal{O}(TDs\lceil \delta N \rceil)$) can be performed in parallel, which reduces the computation time compared to methods that directly deal with the whole dataset. Fig.~\ref{fig:res_syn_d} supports this analysis.

\section{Conclusions}
\label{sec:sec5}
We have proposed a parallelizable multi-subset based self-expressive model for subspace clustering.
A representation of the input data is formulated by combining the solutions of small optimization problems with respect to multiple subsets generated by data sampling.
We have shown that this strategy can significantly improve speed with a multi-core approach that can be easily implemented, especially for large-scale data.
Moreover, it has been verified that combining multiple subsets can reduce the self-expressive residuals of data compared to a single subset.
Extensive experiments on synthetic data and real-world datasets have demonstrated the efficiency and effectiveness of our approach. As a limitation, our method is still unable to handle nonlinear subspaces due to the problem setting. In  future, we would like to design a self-expressive model that can handle nonlinear subspaces, with the help of modeling capabilities from neural network architectures.

\bibliographystyle{IEEEtran}
\bibliography{myref}
%



\end{document}